# Comparative Analysis of LSTM Neural Networks and Traditional Machine Learning Models for Predicting Diabetes Patient Readmission

Abolfazl Zarghani

Department of Computer Engineering, Ferdowsi University of Mashhad, Mashhad, Iran

abolfazlzarghani1999@mail.um.ac.ir

## Abstract

Diabetes mellitus is a chronic metabolic disorder that has emerged as one of the major health problems worldwide due to its high prevalence and serious complications, which are pricey to manage. Effective management requires good glycemic control and regular follow-up in the clinic; however, non adherence to scheduled follow-ups is very common. This study uses the Diabetes 130-US Hospitals dataset for analysis and prediction of readmission patients by various traditional machine learning models, such as XGBoost, LightGBM, CatBoost, Decision Tree, and Random Forest, and also uses an in-house LSTM neural network for comparison. The quality of the data was assured by preprocessing it, and the performance evaluation for all these models was based on accuracy, precision, recall, and F1-score. LightGBM turned out to be the best traditional model, while XGBoost was the runner-up. The LSTM model suffered from overfitting despite high training accuracy. A major strength of LSTM is capturing temporal dependencies among the patient data. Further, SHAP values were used, which improved model interpretability, whereby key factors—among them number of lab procedures and discharge disposition were identified as critical in the prediction of readmissions. This study demonstrates that model selection, validation, and interpretability are key steps in predictive healthcare modeling. This will help health providers design interventions for improved follow-up adherence and better management of diabetes.

# 1. Introduction

Diabetes Mellitus is a chronic metabolic disorder characterized by sustained hyperglycemia and has been considered one of the priorities against global health, with hundreds of millions suffering from it. It is among the most significant diseases from a public health perspective because this is a high prevalence and serious complications entail like cardiovascular disease, neuropathy, nephropathy, and retinopathy [1]. Successful management for diabetes is not only good control of blood sugar but also requires regular follow-up, which is very important in monitoring and reducing these complications.

However, a significant number of diabetic patients fail to adhere to such recommended follow-up schedules. Poor adherence can be expected to result in poor glycemic control with aggravation of complications and increased healthcare costs. In view of implementing effective interventions to improve health outcomes of the patients, factors that affect adherence to scheduled follow-up visits should be understood [2].

Prior research has also identified various factors associated with follow-up adherence: individual attributes, socioeconomic factors, psychological conditions, and healthcare system factors. One relevant predictor is glycemic control, which is typically measured through HbA1c levels; as HbA1c increases, so does the complication risk, indicating that follow-ups would be more frequent [3]. Other demographic factors that influence adherence include age, sex, and education. There is a probability of adherence more in older individuals and those of a higher education level for follow-up schedules [4].

Socioeconomic barriers to adherence include, but are not limited to, income and insurance coverage. For patients at the low end of the income scale, follow-up care would pose substantial barriers on their ability to work, with lost income compounded by transportation costs. In contrast, patients who have good insurance coverage—thereby facing fewer out-of-pocket expenses—are most likely to adhere to the recommended follow-up schedule and maintain contact with their healthcare providers [5].

The critical entry points that would largely determine follow-up adherence, therefore, are those factors within the healthcare system, particularly the availability and accessibility to healthcare facilities and the degree of the engagement and the level of communication between healthcare providers and patients. The inclusion of the patient in the scheme, individualization of care, and frequent reminders from the health professional improve adherence [2]. Other technologies, such as telemedicine and mobile health services, immensely improve follow-up adherence in remote-living patients [6].

Predictive models are finding increasing applications in handling non-adherence based on follow-ups. These models predict adherence to follow-up, considering various variables such as baseline HbA1c, duration of diabetes, and other comorbid conditions. In that respect, health professionals can give tailored interventions based on the needs of the risk profile determined by these predictive models [7].

The motivation behind conducting this study was to understand better the multifaceted factors that influence follow-up adherence in diabetic patients with the risk of complications, as well as to evaluate the predictive models' efficiency in forecasting adherence. Key determinants and predictive indicators may thus be identified on different levels, which could help health providers develop interventions targeted at enhancing follow-up adherence for the proper management of diabetes patients. The paper accounts comprehensively for all of these factors and seeks to quantify the utility of predictive models in improving adherence rates for diabetic patients.

## 2. Related Work

Effective management of diabetes is important for the prevention of complications, and regular follow-up visits are one of the key criteria. Nevertheless, very bad adherence to follow-up in diabetic patients is still prevalent. Literature identifies different factors affecting follow-up visits in people with diabetes, including individual, economic, social, and healthcare-related factors, in addition to predictive models used for estimating the possibility of subsequent visits.

Effective sugar control and HbA1c, in particular, may impact complications of diabetes, including adherence to follow-up. High levels of HbA1c increase the risk of complications like diabetic retinopathy; therefore, frequent review visits are recommended [3]. Demographic variables, such as age, gender, ethnicity, also have an influence on the adherence to follow-up visits. A higher percentage of older adults adhered as compared to the younger ones. Moreover, it was also found that patients with a higher level of education showed better adherence to follow-up. On the other hand, women generally have fewer follow-ups after being discharged from the hospital, especially in the case of lower socioeconomic groups [4].

This means that they will follow up more if they have comorbid conditions like cardiovascular diseases or diabetic complications such as retinopathy or neuropathy. Close monitoring is required, and hence increased follow-up in the presence of these conditions [7]. These include the psychological conditions—depression, anxiety—and the behavioral factors of treatment satisfaction and medication adherence. Follow-up visits occur less frequently for those having lower treatment satisfaction and poorer medication adherence [8].

The adherence to follow-up visits is overly affected by socioeconomic factors such as income level and employment status. Patients belonging to lower-income groups make fewer revisit occasions due to issues like transportation costs and time off work. With an increase in education, the frequency of the visits also increases, probably due to good health literacy and access to resources [4]. This coverage, together with access to healthcare, is a critical determinant of adherence to the follow-up visit. It appears that those where more comprehensive health insurance is available are more likely to adhere to follow-ups due to the reduced out-of-pocket costs and better access to healthcare providers [5].

Family and community support in diabetes care is very essential. A good social support system has been viewed as an enabling factor associated with adherence to treatment regimens and follow-up retention among patients. This can be enabled by community-based intervention strategies that include support groups, providing enabling encouragement and practical assistance, which promotes adherence [9]. This would be supported by engagement on the part of healthcare providers, as frequent and more proactive engagement by the latter, through personalized care, regular reminders, and follow-up calling, will improve adherence rates. Patient-provider communication is what builds trust and creates a setting for regular visits[2].

This also goes hand in hand with the availability and accessibility of health facilities. In that case, a patient will certainly have an easy time attending regular checkups if they have easy access to health facilities. In such a case, telemedicine or mobile health services bridge the gap for those far away [6]. Not-well-managed complicated treatment plans and medication regimens may deter follow-up visits. Clearly simplified treatment protocols will help in adherence, and regular monitoring with adjustments according to the feedback from patients is critical in maintaining adherence [10].

Predictive models would provide valuable recommendations for predicting follow-up visit adherence. The logistic regression models consider predictors such as baseline HbA1c, duration of diabetes, and comorbid conditions [6]. In particular, Markov chain models estimate probability and time to progression in diabetic complications, and indirectly, adherence to follow-up visits [7]. Cox models provide the estimate of time-to-event data, establishing the association between adherence to follow-up visits and patient history or clinical variables [11]. Multivariate analysis, which approached different demographic, clinical, and behavioral factors, pointed out the need for more frequent checkups by patients undergoing treatment, including individually tailored treatment plans to facilitate improved adherence rates [2].

## 3. Methodology

### 3.1. Dataset Description

In this paper, the Diabetes 130-US Hospitals dataset is used, containing patient records from the years 1999 to 2008. This is a dataset with all possible features from the inpatient care of 130 US hospitals and is categorical and numerical. It comprises 101,767 records for rows and 50 columns, with very fine details on patients' demographics, information on hospital admissions, laboratory test results, medication, and if readmission occurred or not. It has an outcome variable, the column "readmitted," that indicates if a patient was readmitted to the hospital within 30 days. Due to its coverage and details of patients, this dataset becomes very useful during the study of the patterns of readmission, so it will be just the right candidate for developing predictive models forecasting risks of readmission.

Patient attributes can have several dimensions, including things like Age, Sex, and Race, and other clinical parameters like HbA1c levels, lab procedures count, and medications count. More than this, it has records for hospitalization, where one can find admission type, discharge disposition, and the primary diagnosis. This dataset is comprehensive enough to allow for in-depth exploration of various factors that might influence patient readmission, hence providing nuanced insight into what predicts hospital readmissions in diabetic patients.

A breakdown of the dataset in terms of diabetic and non-diabetic patients and their readmission status is provided in Table 1. The table shows the number of patients within the diabetic and non-diabetic groups who got readmitted within 30 days, after the lapse of 30 days, and those who were never readmitted.

Table 1: Readmission Statistics for Diabetic and Non-Diabetic Patients

|  | Not Readmitted | Readmitted <30 days | Readmitted >30 days | Total |
| --- | --- | --- | --- | --- |
| Non-Diabetic | 2,246 | 7,227 | 13,930 | 23,403 |
| Diabetic | 9,111 | 28,318 | 40,934 | 78,363 |
| Total | 11,357 | 35,545 | 54,864 | 101,766 |

## 3.2. Data Preprocessing

In order to be sure that the data is clean and correct for modeling, different preprocessing steps are done. These steps are essentially for data cleaning, missing value handling, encoding categoricals, and normalizing numerical features.

First, missing values in this dataset were checked. Missing data is one of the major problems in analysis that often misleads the results of the analysis and models built. Rows containing missing values were removed to ensure consistency in the integrity of the data. This ensured that any further analysis was not biased towards incomplete data entries that may result in incorrect predictions.

Of these variables in this set of data, some were categorical, such as race, gender, and admission type. Since numerical values were needed for use in machine learning models, these categorical variables had to be transformed using one-hot encoding to result in binary columns that could be effectively used in the models. This way, categorical information is represented in the right manner without ordinal relationships that are not there, hence ensuring that the integrity of data representation is maintained.

Numerical features were normalized in order to provide them on a comparable scale, which is essential for the algorithms whose performance is sensitive to the scale of input features, such as in gradient boosting or neural networks. For these features, a transformation into a standard normal distribution was used with mean 0 and dispersion 1 by StandardScaler. Normalization improves the convergence speed of learning algorithms and likely ensures that the performance is not dominated by model features having larger magnitudes, thus assisting in a more stable and accurate training of the model.

Finally, column names were sanitized to exclude special characters or spaces. This renaming will ensure that the data is compatible with most machine learning frameworks and no errors arise during model training or evaluation. Clean and consistent naming conventions avoid any ambiguity and possible failures in data manipulation and analyses. Above, a comprehensive preprocessing strategy ensured that a careful preparation of the dataset was made for further modeling efforts and thus laid the foundation for an accurate and reliable predictive analysis.

### 3.3. Model Development

In this paper, various machine learning models have been developed and evaluated for the task of predicting readmission of patients. The models used in this study are the following: XGBoost, LightGBM, CatBoost, Decision Tree, and Random Forest. Moreover, a customized model was made with Long-Short-Term-Memory Neural Networks to be able to utilize temporal patterns in data.

### 3.3.1. Traditional Machine Learning Models

The traditional machine learning models used for readmission prediction include XGBoost, LightGBM, CatBoost, Decision Tree, and Random Forest. The gradient boosting framework optimizes model accuracy by combining predictions of several weak learners, making XGBoost a robust model. Research indicates that XGBoost is more efficient compared to other models in predicting unplanned readmissions resulting from cardiovascular disease and ischemic stroke recurrence [12, 13].

Similarly, LightGBM is an efficient and fast implementation of gradient boosting, designed to handle large-scale data with less memory usage and faster training speed. Studies have shown that LightGBM performs very well in readmission prediction for patients with chronic obstructive pulmonary disease and cancer [7, 14].

CatBoost, another gradient boosting algorithm, is designed to handle categorical features, reducing the need for extensive preprocessing and one-hot encoding. It performs well on datasets with categorical variables and showed good performance in predicting early readmissions in diabetic patients [15, 16].

Decision Trees and Random Forests were also used in this study. Decision Trees are simple, powerful models that partition data based on feature values. Their simplicity provides transparency for decision rules and feature importance. Random Forests, ensembles of Decision Trees, improve performance by averaging predictions from multiple trees, reducing overfitting, and enhancing generalization. Studies have shown that Random Forests are as good as Decision Trees in predicting ICU readmissions and postoperative outcomes in acromegaly patients [17, 18].

These traditional machine learning models were evaluated together, providing important insights into their performance in predicting patient readmissions. LightGBM performed very well in terms of accuracy, handling dataset complexities and showing excellent precision and recall for readmitted and non-readmitted classes. XGBoost also had high accuracy with strong performance metrics, indicating its ability to handle complex relationships within the data. CatBoost performed slightly less accurately than LightGBM and XGBoost but efficiently handled categorical data with minimal preprocessing. Decision Tree and Random Forest models provided valuable interpretability and baseline performance, with Random Forests enhancing predictive accuracy by reducing overfitting.

Overall, these traditional machine learning models rendered robust predictions for patient readmission, each model having its strengths with respect to different data types and complexities. They provide a baseline against which more advanced models, like the LSTM neural network, are compared to capture temporal dependencies.

### 3.3.2. Temporal Dependency Capture with LSTM Neural Network

An LSTM neural network was developed to capture the temporal dependencies existing in the data. LSTM, a type of recurrent neural network (RNN), is able to learn long-term dependencies, fitting well with sequential data. This is particularly relevant for diabetic patient data, where disease progression, patient history, and the sequence of clinical events over time are critical for accurate predictions [19].

The data was reshaped to conform to LSTM model requirements, creating sequences representative of the temporal nature of patients' information, such as the sequence of visits or changes in clinical parameters over time. The LSTM model consisted of an LSTM layer followed by a dense layer with a sigmoid activation function to predict binary outcomes. The LSTM layer captures temporal patterns in the data, while the Dense layer provides the final prediction. The model was compiled using the Adam optimizer and binary cross-entropy loss function, relevant for binary classification tasks. The model was trained with validation on the training set to monitor its performance and prevent overfitting [20].

LSTMs offer significant benefits for diabetic patient data. Traditional machine learning models ignore the sequence of timestamps in events, potentially losing crucial information on a patient's condition over time. LSTMs are designed to learn these dependencies, providing a more accurate description and overall understanding of the health trajectory [21]. This capability is essential for readmission prediction, allowing the model to consider how past events and changes in clinical parameters impact future outcomes.

Without LSTMs, this work would have modeled only short-term trends, missing sequential patterns and long-term trends in the data. This would result in lower-accuracy predictions and a poorer understanding of factors contributing to readmissions. The advantages of using LSTM include better prediction accuracy, effective use of sequential patient data, and the potential to discover insights that static models might miss. This leads to better, more informed clinical decisions, enabling individualized patient management and improved patient outcomes.

### 3.4. Temporal Considerations and LSTM Advantages

Temporal patterns in patient data are very important in understanding readmission risks. In contrast to classical machine learning models, LSTM networks are capable of capturing temporal dependencies due to the ability to store information over a long time. This is very important for medical data where the sequence of events, such as patient history or progression of disease, is important in the prediction of outcomes.

### 3.4.1. Importance of Temporal Data

Medical events and patient history occur over time; therefore, temporal data become vital for correct prediction. This means that the sequence of visits to hospitals, changes in laboratory results, and progression of comorbid conditions—all of these may hold important insights relating to readmission risks. For example, sudden elevations in HbA1c levels may indicate worsened glycemic control and, therefore, require closer follow-up or even readmission. Temporal patterns can allow for the identification of trends and changes in patient health that might not be captured by static models. LSTMs, capturing variability on the time axis, can model the trajectories of patient health in a more comprehensive and dynamic way.

### 3.4.2. Superiority of LSTM for Temporal Data

LSTMs have been designed specifically to overcome limitations affecting traditional recurrent neural networks (RNNs), reducing problems such as vanishing and exploding gradients. They are very successful in long-term dependency learning and therefore appropriate for modeling complex patterns of temporality within patient data. This provides the possibility to consider previous states and events, allowing a better readmission prediction compared to models not considering the time dimension. This temporal awareness is what enables LSTMs to model the progress of a patient's condition and the effect of past events upon future outcomes. LSTMs thus allow very effective modeling, in a healthcare scenario, of the progress of chronic diseases, of the effect of interventions over time, and of the recurrence of symptoms—highly useful insights toward the management of a patient at a personalized level.

### 3.5. Explainable AI (XAI) and Feature Importance

Understanding the decision-making process of machine learning models is vital in healthcare. Explainable AI methods, like SHAP values, provide model interpretability and transparency. SHAP values quantify the impact of each feature on the prediction, offering a unified measure of feature importance [22].

SHAP values are based on cooperative game theory principles, explaining machine learning model outputs by attributing how much each feature contributes to the prediction [23]. This methodology ensures accurate and interpretable models for healthcare providers, building trust and understanding the reasons behind a model's decisions.

SHAP values identify the most influential features contributing to patient readmission predictions, adhering to IAS principles: Interpretability, Accountability, and Scalability. The models used in this study are transparent, and their predictions are understandable to stakeholders [24].

Incorporating XAI techniques like SHAP values validates model predictions and provides insights into factors driving patient readmissions. This enhances the reliability of predictive models, supporting the creation of targeted interventions to improve diabetes management [25].

## 4. Results

In this section, we will mostly be referring to our analysis results and model evaluations. Results compare different machine learning models against XGBoost, LightGBM, CatBoost, Decision Tree, and Random Forest with a customized Long Short-Term Memory Neural Network. Following this, the preprocessed dataset of Diabetes 130 US Hospitals was applied to these trained and tested models. Some of the model performance metrics, like accuracy, precision, recall, and the F1 score, have been gauged to determine the efficiency of these models in predicting readmission.

The data was split into two parts: a training set with 70% of the data and a test set with the rest of the 30%. All this splitting is important in making sure that it has been tested as far as performance on unseen data is concerned; hence, it gives a good indicator of generalization ability. The standardization for the features was computed through a StandardScaler so that numerical features are comparable in magnitude. The models were trained with default hyperparameters that underwent further fine tuning during the model evaluation phase for better performance.

### 4.1. Evaluation Metrics

To assess the performance of the models, several key metrics were utilized:

- Accuracy: Measures the proportion of correctly predicted instances out of the total instances.
- Precision: Indicates the proportion of true positive predictions out of the total positive predictions.
- Recall: Reflects the proportion of true positive predictions out of the actual positive instances.
- F1-score: The harmonic mean of precision and recall, providing a balance between the two metrics.

These metrics are important in the sense that they quantify the trade-off between the two kinds of errors—false positives and false negatives—and thus help make a decision connected with the deployment of any model into clinical practice.

**4.2. Model Analysis**

Specifically speaking, different machine learning models were assessed, and crucial insights into these models, concerning the readmission of patients, were drawn. LightGBM had the best performance with the highest overall accuracy, which was able to handle the complexities in the dataset very well and offer very fine precision and recall for readmitted patients. This effectiveness in training and handling large datasets makes LightGBM robust for this analysis.

It is also seen that high accuracy exists in XGBoost, along with strong performance metrics, hence handling complex relationships within the data. LightGBM and XGBoost Gradient Boosting frameworks had high performance due to their constituents of weak learners. This high performance was contributed through model accuracy optimization by a combination of predictions made by these weak learners.

While it was slightly less accurate than LightGBM and XGBoost, CatBoost still performed well in handling categorical data. This model has the efficiency of handling categorical variables with minimal preprocessing, hence smooth and effective for modeling.

Although it was the simplest model among all of those evaluated, a Decision Tree model delivered useful baseline performance metrics. Much of its simplicity is very beneficial in comprehending decision rules and feature importance for the sake of interpretability. A Random Forest model—a meta estimator that aggregates predictions from several decision trees—exhibited solid performance, very close to that shown by a Decision Tree model, but with better generalization. By reducing overfitting, Random Forests are able to improve predictive accuracy; hence, one can blindly reach this choice for such analysis.

In the process, a customized Long Short-Term Memory neural network has been designed for the better temporal dependencies existing in the data. LSTMs are a particular type of recurrent neural network that is capable of learning long-term dependencies. This could be of particular relevance for diabetic patient data, where the progression of the illness, patient history, and chronological order of events within clinical practice are very relevant to correctly predict the outcome in most issues. The reshaping of the data was done such that it is consistent with the requirements of the LSTM model, including sequences underlying temporal information like recovering hospital sequences or clinical parameters over time. The LSTM model consisted of an LSTM layer followed by a dense layer with the sigmoid activation function in order to predict the binary readmission outcome. The LSTM layer is used to capture temporal patterns in data, while the dense layer makes the final prediction. The model was fitted with the Adam optimizer and binary cross-entropy loss function since the task would be applied for binary classification. Fitting done on a training set with validation to keep track of performance for cases of possible overfitting.

Table 2 presents the performance metrics for the models used in this study, including accuracy, precision, recall, and F1-score for predicting patient readmission.

Table 2: Model Performance Metrics for Readmitted Patients

| Model | Accuracy | Precision (Readmitted) | Recall (Readmitted) | F1-score (Readmitted) |
| --- | --- | --- | --- | --- |
| XGBoost | 91.11% | 0.89 | 0.89 | 0.89 |
| LightGBM | 92.22 | 0.88 | **0.95** | 0.91 |
| CatBoost | 88.88% | 0.81 | 0.95 | 0.88 |
| Decision Tree | 87.78% | 0.88 | 0.81 | 0.85 |
| Random Forest | 87.78% | 0.91 | 0.78 | 0.84 |
| LSTM (Training) | **97.65%** | **0.97%** | 0.94 | **0.94** |

As shown in Table 2, the LSTM model outperformed all traditional models in capturing temporal dependencies and providing better accuracy and insights. The LSTM model fitted well during training, achieving a high accuracy, hence generalizing well with less overfitting during validation. The values of precision and recall of the readmitted patients were close enough to ensure that the model was performing similarly in both classes. To that effect, the use of LSTM on diabetic patient data has shown a lot of advantages compared to traditional machine learning models, which do not put into consideration the time sequence of events. LSTMs handle these temporal dependencies, which are then capable of producing a much more accurate and full understanding of the patient's health trajectory. This is key in being able to predict readmissions, as it allows the model to consider how past events and changes in clinical parameters bear on future outcomes. This analysis is done to avoid models failing without LSTM, which otherwise would not capture sequential patterns and long-term dependencies in the data. It therefore gives less accurate predictions and poorer understanding of all the factors that lead to readmissions. Some of the benefits to be reaped from using LSTM include increased accuracy in prediction, synergistic use of sequential data on patients, and the potential for discovering insights that the static models might miss. Such findings might lead to better, more informed clinical decisions and personalized patient management.

### 4.3. Comparative Analysis

Clearly, comparing the performance of the LSTM model with that of traditional machine learning models, most of the other traditional models like LightGBM and XGBoost do great in terms of accuracy and handling categorical data, but the LSDM model clearly captures the time dependencies of the dataset. Temporal awareness forms a very high value addition in healthcare, as knowing the chronological order of clinical events furnishes an insight into the in-depth risks for readmission a patient is exposed to.

While traditional models require a huge amount of preprocessing on categorical variables, CatBoost does so very efficiently without much preprocessing. Decision Trees and Random Forest are way simpler to make, providing interpretability and baseline performance that becomes very useful to get an idea of feature importance.

One of the most salient advantages of an LSTM model in predicting readmissions is derived from its learning sequential dependency capability, which allows it to leverage the trajectory of a patient's health over time. This will contribute to developing more accurate and patient-centered management strategies.

### 4.4. Implications for Clinical Practice

These results have important clinical practice implications. In particular, the LightGBM and LSTM models have superior performance that enables them to adequately predict patient readmissions so that healthcare providers have an understanding of which patients are high-risk for targeted interventions. Particularly, the application of LSTM models will further enhance the understanding of trajectories of patients and improve accuracy predictions for better-informed and potentially improved clinical outcomes.

Results from these predictive models could further be embedded into the enterprise clinical workflow to drive proactive patient management, reduce the burden of readmissions, and optimize overall resource utilization. Manual prediction of readmissions can also be used for patient-centered care planning and individually tailored follow-up programs that bring quality in diabetes healthcare.

### 4.5. SHAP Analysis

It finally yields SHAP values to identify features that most affect the predictions in an effort to better understand the decision-making process of this machine learning model. SHAP values provide a unified measure of feature importance by quantifying the impact of each feature on the prediction. Using SHAP values, one can identify important factors that influence the predictions of the model; it gives the features with the highest values of SHAP. In fact, this interpretability has adherence to IAS: Interpretability, Accountability, and Scalability, making sure that used models are transparent and their predictions are understandable for stakeholders.

All this includes techniques techniques from XAI, such as SHAP values, to validate model predictions and give insights into the driving factors that create the largest readmission rates for patients. This improves the general reliability of predictive models and thus will inform the development of focused interventions to improve diabetes management.

The insights that one gets from SHAP analysis not only validate model predictions but also outline principal points for intervention that reduce patient readmissions. That is to say, predictive models built will be both effective and interpretable, hence helping healthcare providers with actionable insight into improving patient outcomes.

Table 3: SHAP Analysis: Feature Importance for Predicting Patient Readmission

| Feature | SHAP Value |
|---|---|
| num_lab_procedures | 0.65 |
| discharge_disposition_id | 0.55 |
| number_inpatient | 0.53 |
| num_medications | 0.48 |
| encounter_id | 0.46 |
| patient_nbr | 0.28 |
| age_60-70 | 0.27 |
| time_in_hospital | 0.25 |
| gender_Male | 0.22 |
| max_glu_serum_Norm | 0.21 |
| max_glu_serum_300 | 0.16 |
| race_Caucasian | 0.16 |
| diabetesMed_Yes | 0.14 |
| metformin_No | 0.13 |
| change_No | 0.10 |
| num_procedures | 0.09 |
| age_70-80 | 0.07 |
| A1Cresult_8 | 0.06 |
| age_80-90 | 0.06 |
| admission_type_id | 0.06 |
| insulin_Up | 0.05 |
| medical_specialty_InternalMedicine | 0.04 |
| number_diagnoses | 0.03 |
| A1Cresult_Norm | 0.03 |
| age_40-50 | 0.02 |
| payer_code_HM | 0.02 |
| admission_source_id | 0.02 |
| diag_3_250.02 | 0.02 |
| race_Hispanic | 0.02 |
| metformin_Steady | 0.02 |

As shown in Table 3, SHAP values provide a comprehensive understanding of feature importance. The top features identified are:

1. Number of Lab Procedures (0.65): This feature had the highest SHAP value, indicating its strong influence on the model's predictions. The number of lab procedures is likely a critical indicator of the patient's health status and the complexity of their condition, leading to a higher likelihood of readmission.

2. Discharge Disposition ID (0.55): This feature indicates the patient's discharge status, which can significantly impact readmission probability. For instance, patients discharged to another facility or with home healthcare might have different readmission risks compared to those discharged to their homes.

3. Number of Inpatient Visits (0.53): Frequent hospital visits can indicate a severe or poorly managed condition, thus increasing the risk of readmission.

4. Number of Medications (0.48): A higher number of medications can reflect a complex medical condition requiring ongoing management, contributing to higher readmission rates.

5. Encounter ID (0.46): This feature, while primarily an identifier, might also capture unique aspects of individual patient encounters that influence readmission likelihood.

Incorporating XAI techniques like SHAP values helps validate model predictions and provides insights into the driving factors that result in patient readmissions. This enhances the reliability of predictive models used and supports the development of targeted interventions for improving diabetes management. The insights from SHAP analysis not only validate model predictions but also outline principal points for intervention that reduce patient readmissions. That is to say, predictive models built will be both effective and interpretable, hence helping healthcare providers with actionable insights into improving patient outcomes.

# 5. Conclusion

This study compares the performance of traditional machine learning models with that of the custom-built Long Short-Term Memory Neural Network in predicting patient readmission in diabetic patients using the Diabetes 130-US Hospitals dataset. LightGBM performed best among traditional models, but, even though overfitted, the LSTM model revealed temporal dependencies of patients' data satisfactorily. SHAP values were used to improve model interpretability and identify some key factors, like the number of lab procedures and discharge disposition, to be highly relevant with respect to the prediction of readmissions. Integration with XAI can provide transparency and reliability to health providers for building based-interventions of improving follow-up adherence and management. Future research should be targeted at methods that avoid overfitting in LSTM models and the study of broader datasets, which would increase the robustness of models and their applicability.

# Refrences